\documentclass{article}

\usepackage{PRIMEarxiv}

\usepackage[utf8]{inputenc} %
\usepackage[T1]{fontenc}    %
\usepackage{hyperref}       %
\usepackage{url}            %
\usepackage{booktabs}       %
\usepackage{amsfonts}       %
\usepackage{nicefrac}       %
\usepackage{microtype}      %
\usepackage{lipsum}
\usepackage{fancyhdr}       %
\usepackage{graphicx}       %
\graphicspath{{media/}}     %

\usepackage{amsmath} %
\usepackage{bbm}
\usepackage{cleveref}
\usepackage{comment} %
\usepackage[numbers]{natbib} %
\usepackage{graphicx} %
\usepackage{verbatim} %

\usepackage{mathtools} %

\DeclareMathOperator*{\argmin}{arg\,min}

\newcommand{\xv}{\mathbf{x}}
\newcommand{\Xv}{\mathbf{X}}
\renewcommand{\xi}{\xv_i}
\newcommand{\fh}{\hat{f}}
\newcommand{\R}{\mathbb{R}}
\newcommand{\Xspace}{\mathcal{X}}
\newcommand{\D}{D} %

\usepackage{framed} %
\usepackage{amsfonts} %
\usepackage{amssymb} %
\usepackage{enumerate} %
\usepackage{color} %
\usepackage{eurosym} %
\usepackage[misc]{ifsym} %
\usepackage{caption}
\usepackage{subcaption}

\usepackage{algorithm}
\usepackage{algorithmic}

\usepackage{enumitem}

\makeatletter
\newcommand{\pfnsymbol}[1]{%
  \textsuperscript{\@fnsymbol{#1}}%
}
\makeatother

\title{CountARFactuals -- Generating plausible model-agnostic counterfactual explanations with adversarial random forests
}

\author{
  Susanne Dandl\thanks{Equal contribution as first authors.}$^{\;\,,1}$, Kristin Blesch\pfnsymbol{1}$^{,2,3}$, Timo Freiesleben\pfnsymbol{1}$^{,5}$, Gunnar König\pfnsymbol{1}$^{,6}$,\\
\textbf{Jan Kapar$^{2,3}$, Bernd Bischl$^{1}$, and Marvin N. Wright $^{3,4,5}$}
\and
$^1$Munich Center for Machine Learning (MCML) and Department of Statistics, LMU Munich
\and
$^2$Leibniz Institute for Prevention Research \& Epidemiology – BIPS  
\and
$^3$Faculty of Mathematics and Computer Science, University of Bremen
\and
$^4$Department of Public Health, University of Copenhagen
\and
$^5$Cluster: Machine Learning for Science, University of Tübingen
\and
$^6$Tübingen AI Center and University of Tübingen\\
\texttt{wright@leibniz-bips.de} \\
}

\begin{document}
\maketitle

\begin{abstract}
Counterfactual explanations elucidate algorithmic decisions by pointing to scenarios that would have led to an alternative, desired outcome.
Giving insight into the model's behavior, they hint users towards possible actions and give grounds for contesting decisions. 
As a crucial factor in achieving these goals, counterfactuals must be plausible, i.e., describing realistic alternative scenarios within the data manifold. 
This paper leverages a recently developed generative modeling technique -- adversarial random forests (ARFs) -- to efficiently generate plausible counterfactuals in a model-agnostic way.
ARFs can serve as a plausibility measure or directly generate counterfactual explanations.
Our ARF-based approach surpasses the limitations of existing methods that aim to generate plausible counterfactual explanations: It is easy to train and computationally highly efficient, handles continuous and categorical data naturally, and allows integrating additional desiderata such as sparsity in a straightforward manner.
\end{abstract}


\section{Introduction}
\label{sec:intro}
Machine learning (ML) algorithms are increasingly used in high-stakes scenarios. For example, they help to decide whether you receive a loan, if you are suitable for a job, or even which disease you are diagnosed with. While ML-based systems are powerful at detecting complex patterns in data, the reasoning behind their predictions is often not easy to discern for humans. 
Many ML models are black boxes with a complex mathematical structure that do not follow transparent logical rules \citep{burrell2016machine}.

The emerging field of \emph{interpretable machine learning} (IML) (also known as explainable artificial intelligence or XAI for short) promises to open up these black boxes and aims to make the decisions of ML models transparent to humans (see \citep{adadi2018peeking,molnar2022} for overviews).
A particularly simple approach is to explain algorithmic decisions to end-users via so-called \emph{counterfactual explanations} \citep{wachter2017counterfactual}. 
\begin{itemize}
    \item[] \textbf{Example:} Imagine you apply for a loan. You enter characteristics such as your age, salary, loan amount, etc. in the online application form and after a few seconds you receive the decision -- your loan application has been denied. A counterfactual explanation could be: If your salary had been \euro 5,000  higher, your loan would have been approved. 
\end{itemize}
More generally, a counterfactual explanation points to a close alternative scenario (the so-called \emph{counterfactual}) that, in contrast to the actual scenario, would have resulted in the desired outcome. Counterfactual explanations may be employed for various purposes, such as helping to guide a person's actions \citep{karimi2022survey,konig2023improvement}, enabling them to contest adverse decisions \citep{lyons2021conceptualising}, and providing insights into the decision behavior of the model \citep{mittelstadt2019explaining}. For all these goals, counterfactuals must be \emph{plausible}, which means the alternative scenarios they depict are realistic. For instance, in the example above, suggesting a negative loan amount or a real estate loan with an amount of \euro 500 would not be very plausible counterfactuals.

When adding plausibility as another objective for generating counterfactuals, its trade-off with proximity, i.e., that the counterfactual is close to the point of interest, should be taken into account. 
Dandl et al. \cite{dandl2020multi} were one of the first to address this trade-off by framing the counterfactual search as a multi-objective optimization problem. Their approach -- multi-objective counterfactual explanations (MOC) -- returns not just a single counterfactual, but a Pareto set of counterfactuals, which is advisable to account for the Rashomon effect, i.e., that multiple, diverse, equally good counterfactuals may exist \citep{breiman2001twocult}.
 
An intuitive approach to plausibility is searching for only those counterfactuals that are close to actual instances in the dataset \cite{guidotti2022counterfactual}. 
To operationalize this goal, one objective in MOC minimizes the distance between counterfactuals and the actual instances.
However, as presented in \Cref{subsec:counterfac}, this approach has its limitations if, for example, there are low-density gaps close to $\xv^*$ between high-density regions. 
Other approaches model plausibility via the joint probability density. They rely on computationally intensive neural network architectures such as variational autoencoders (VAEs) \citep{brughmans2023nice,joshi2019towards,mahajan2019preserving,pawelczyk2020learning} or generative adversarial networks (GANs) \citep{nemirovsky2022countergan,van2021conditional}. While these architectures have merits for high-dimensional tensor data (e.g., images or text), they are less suited for tabular data (see our discussion in \Cref{subsec:arfs}).

\paragraph{Contributions}
We leverage a tree-based technique from generative modeling called \emph{adversarial random forests} (ARF) \citep{watson2023} to generate plausible counterfactuals in a mixed (i.e., categorical and continuous) tabular data setting. 
We call these countARFactuals and  propose two model-agnostic algorithms to generate them:
\begin{enumerate}
    \item We integrate ARF into the multi-objective counterfactual explanation (MOC) framework \cite{dandl2020multi} to speed up the counterfactual search and find more plausible counterfactuals (see \Cref{subsec:algo1-MOC+ARF}).
    \item We tailor ARF to directly generate plausible counterfactuals without an optimization algorithm (see \Cref{subsec:algo2-onlyARF}). 
\end{enumerate}
A simulation study shows the advantages in plausibility and efficiency of our ARF-based approaches compared to competing methods (\Cref{sec:experiments}).
Moreover, we apply our method on a real-world dataset, namely to explain coffee quality predictions (\Cref{sec:real-data-example}).

\section{Related Work}
\label{sec:related-work}
There is widespread agreement in the counterfactual community that plausibility is an important concern \citep{freiesleben2022intriguing,guidotti2022counterfactual,karimi2022survey,keane2021if,stepin2021survey,verma2020counterfactual}.  Various suggestions have been made to incorporate plausibility into the counterfactual search, for example using causal knowledge \citep{konig2023improvement,mahajan2019preserving}, case-based reasoning \citep{keane2020good}, outlier detectors \citep{kanamori2020dace}, restricting the search space \citep{artelt2020convex}, imputing feature combinations from real instances \citep{goyal2019counterfactual}, respecting paths between datapoints \citep{poyiadzi2020face}, or, as described above, staying close to the training data \citep{dandl2020multi}.

Many define plausibility theoretically through the joint probability density \citep{verma2020counterfactual}. Some works rely on VAEs or standard autoencoders: they directly generate counterfactuals  \cite{mahajan2019preserving,pawelczyk2020learning}, use VAEs in the optimization \cite{joshi2019towards} or just for measuring plausibility
\citep{brughmans2023nice}. Other works rely on GANs to generate counterfactuals \citep{van2021conditional,nemirovsky2022countergan}.  However, these approaches differ substantially from our work, as they are tailored for neural network models \citep{mahajan2019preserving}, focus only on plausibility thereby ignoring other objectives like sparsity \citep{mahajan2019preserving,pawelczyk2020learning} (see \Cref{subsec:counterfac}), or work only for continuous data \citep{joshi2019towards,nemirovsky2022countergan}. 
The closest works to ours are Brughmans et al. \cite{brughmans2023nice} and Dandl et al. \cite{dandl2020multi}. Both are designed to generate plausible and sparse counterfactuals in mixed tabular data settings. Brughmans et al. \cite{brughmans2023nice} use the autoencoder reconstruction loss as a plausibility measure and Dandl et al. \cite{dandl2020multi} use the distance to the $k-$nearest neighbors to evaluate plausibility. We show in our experiments in \Cref{sec:experiments} that utilizing ARF to generate counterfactuals improves plausibility compared to those approaches while being computationally fast. %

\section{Background}
\label{sec:background}
Before we present our approaches, we provide background on the two methods we build upon: multi-objective counterfactual explanations (MOC) \cite{dandl2020multi} and adversarial random forests (ARF) \cite{watson2023}.

We consider a supervised learning setup with a binary classification or regression problem.\footnote{Our framework also generalizes to multi-class problems; we restrict ourselves here only for the sake of simplicity and notation.} 
$\mathcal{X}$ denotes a $p$-dimensional feature space.
The respective vector $\Xv:=(X_1,\dots,X_p)^T$ of random variables may contain both continuous and categorical features.
With $Y \in \R$, we denote a random variable reflecting the outcome.
In case of a binary classification model, we restrict $Y$ to $\{0, 1\}$.

To predict $Y$ from $\Xv$, we trained an ML model $\hat{f}:\mathcal{X} \rightarrow \R$ on a dataset $D_{\text{train}}:=\lbrace (\xv^{(1)},y^{(1)}),\dots, (\xv^{(n_{\text{train}})},y^{(n_{\text{train}})})\rbrace$ with $n_{\text{train}}$ observations.
For binary classification, the model output is restricted to $\hat{f}(\xv)\in[0,1]$, reflecting the probability for $Y = 1$. 
Most counterfactual explanation methods require access to a dataset for generating counterfactuals. To reflect that this dataset \textit{can} differ to $D_{\text{train}}$, we denote it as $\D$ in the following and assume it to consist of $n$ observations.

\subsection{Multi-objective counterfactual explanations}
\label{subsec:counterfac}
\begin{figure}[ht]
\centering
    \begin{subfigure}[b]{0.45\textwidth}
         \centering
         \includegraphics[width=0.7\textwidth]{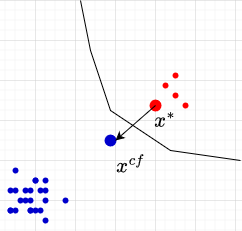}
        \caption{Plausibility-proximity trade-off}
        \label{fig:plaus-prox-tradeoff}
     \end{subfigure}
     \begin{subfigure}[b]{0.45\textwidth}
         \centering
         \includegraphics[width=0.7\textwidth]{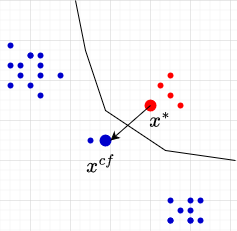}
         \caption{Limitation of MOC's plausibility}
         \label{fig:plausibilityFailMoc}
     \end{subfigure}
     \caption{(a) Proximity and plausibility can be conflicting objectives \citep{dandl2020multi}; enforcing proximity may lead to unrealistic counterfactuals. (b) To have high proximity (i.e., low $o_{\text{prox}}$ in \Cref{eq:moc-prox}) and high plausibility (i.e., low $o_{\text{plaus}}$ in \Cref{eq:moc-plaus}, with $k = 1$), the counterfactual may be in a low-density region.}
\end{figure}

Suppose we want to explain why a certain data point of interest $\xv^*$ was predicted as $\hat{f}(\xv^*)$ instead of a desired prediction within $Y_{des}\subset \R$. Wachter et al. \cite{wachter2017counterfactual} define counterfactuals as the closest possible input vector $\xv^{cf}$ to $\xv^*$ according to some distance on $\mathcal{X}$ such that $\hat{f}(\xv^{cf})\in Y_{des}$. 
This definition does not explicitly demand sparse or plausible changes. 
When integrating all these desiderata into an objective to generate counterfactuals, trade-offs between the different objectives must be taken into account since the objectives conflict each other. 
\Cref{fig:plaus-prox-tradeoff} illustrates this for the properties plausibility and proximity to the original instance $\xv^*$. If all high-density regions are far away from the decision boundary, enforcing proximity leads to unrealistic counterfactuals. 

To consider these trade-offs, Dandl et al. \cite{dandl2020multi} turned the search for counterfactuals into a multi-objective optimization problem:
 \begin{equation}
    \xv^{cf}\in \underset{\xv\in\mathcal{X}}{\argmin} \left( o_{\text{valid}}(\fh(\xv),Y_{des}),o_{\text{prox}}(\xv,\xv^*),o_{\text{plaus}}(\xv,\D),o_{\text{sparse}}(\xv,\xv^*)\right). 
    \label{eq:mutli-objective}
\end{equation}
The different objectives denote:
 \begin{enumerate}
    \item \textbf{Validity}: Counterfactuals should have a predicted outcome in $Y_{des}$
\begin{equation*}
    o_{\text{valid}}(\hat{f}(\xv),Y_{des}):=\underset{y\in Y_{des}}{\inf}
            |\hat{f}(\xv)-y|.
\end{equation*}
    \item \textbf{Proximity}: Counterfactuals should be close to $\xv^*$ according to the Gower distance $d_{\text{Gower}}$ \citep{gower1971}
    \begin{equation}
        o_{\text{prox}}(\xv,\xv^*):=d_{\text{Gower}}(\xv,\xv^*).
        \label{eq:moc-prox}
    \end{equation}
    \item \textbf{Plausibility}: Counterfactuals should describe a realistic data instance, with $\xv^{[1]},\dots,\xv^{[k]}$ indicating the $k-$nearest neighbors to $\xv$ within data $\D$ and $w_i$ denoting weights with $\sum_{i=1}^k w_i=1$
    \begin{equation}
        o_{\text{plaus}}(\xv,\D):=\sum\limits_{i=1}^k w_i d_{\text{Gower}}(\xv,\xv^{[i]}).
        \label{eq:moc-plaus}
    \end{equation}
    \item \textbf{Sparsity}: Counterfactuals should vary from $\xv^*$ in only a few features
    \begin{equation*}
        o_{\text{sparse}}(\xv,\xv^*):=\|\xv-\xv^*\|_{0}=\frac{1}{p}\sum\limits_{j=1}^p \mathbbm{1}_{x_j\neq x_j^*}.
    \end{equation*}
\end{enumerate}
Dandl et al. \cite{dandl2020multi} adapted the nondominated sorting genetic algorithm or short NSGA-II of Deb et al. \citep{deb2002fast} to solve the multi-objective optimization problem. %
This algorithm follows three steps:
\begin{enumerate}
    \item  It generates a set of candidate instances close to the point of interest $\xv^*$. Among these, it recombines and mutates the candidates that perform best according to the above criteria.
    Per default, the mutator does not take feature dependencies into account. 
    To enhance plausibility, mutation can be optionally performed by sampling from conditional distributions learned on $D$ by conditional trees \cite{Hothorn2021} -- we refer to this MOC version as MOCCTREE. 
    \item Both new and old candidates are ranked using nondominated and crowding distance sorting. Nondominated sorting ranks according to optimality with respect to the above objectives (with the option to penalize invalid counterfactuals) and crowding distance ranks according to diversity. 
    \item Based on these rankings, optimal and diverse candidates are selected for the next iteration. The search for counterfactuals ends after either a fixed number of predefined iterations or when the generated counterfactuals are not significantly better according to the hypervolume of the objectives above. As a final step, the algorithm outputs the Pareto optimal set of counterfactuals over the generations.
\end{enumerate}
The conceptualization of plausibility as \Cref{eq:moc-plaus} has its limitations as, e.g., illustrated in \Cref{fig:plausibilityFailMoc}: With $k = 1$ (the default in MOC), counterfactuals with low values in \Cref{eq:moc-plaus} might still end up in low-density regions.

\subsection{Generative modeling and adversarial random forests}
\label{subsec:arfs}
Generative modeling is concerned with models that generate synthetic data $\Tilde{D}$ that mimic the appearance of real data $D$. A well-known approach are VAEs \citep{kingma2013}, which encode original data instances into a set of low-dimensional distribution parameters and then reconstruct these instances with a decoder neural network from samples of these distributions. Another common technique are GANs \citep{goodfellow2014}, where two different neural network models play a zero-sum game -- the generator network aims to generate realistic instances, and the discriminator network aims to discriminate these instances from real data. Other generative models based on neural networks include normalizing flows \citep{rezende2015}, diffusion probabilistic models \citep{ho2020} and transformer-based models \citep{vaswani2017} (see \cite{bond2021,foster2022} for overviews). While there exist adaptions of neural network models to tabular data, tree-based approaches may be better suited \citep{borisov2022, grinsztajn2022, shwartz2022}.

ARFs are a tree-based procedure for generative modeling \citep{watson2023}. The ARF approach is similar to the approach of GANs, however, instead of a neural network as a base learner, ARF relies on random forests. An ARF is trained in three steps: (1) Fitting an unsupervised random forest \citep{shi2006}, which generates a naive synthetic dataset $\Tilde{D}_{1}$ and subsequently trains a random forest $\hat{g}_1$ to distinguish between $D$ and $\Tilde{D}_{1}$. (2) Sampling feature values marginally from the instances in the leaves of $\hat{g}_1$ to obtain a more realistic synthetic dataset $\Tilde{D}_{2}$. Another random forest $\hat{g}_2$ is trained to distinguish between $D$ and $\Tilde{D}_{2}$. (3) This process is repeated until the random forest classifier can no longer distinguish synthetic from real data.
We denote the final ARF model as $\hat{g}^*$.
As opposed to GANs, ARFs allow for both density estimation and generative modeling.
The two algorithms are called \emph{forests for density estimation} (FORDE) and \emph{forests for generative modeling} (FORGE), respectively.

\paragraph{Density estimation with FORDE}
leverages the mutual independence across features in the leaves after algorithm convergence, which allows to model the joint density $p(x)$ as a mixture of univariate feature densities:  
    \begin{equation}
        \text{FORDE}(\xv)\coloneqq\hat{p}(\xv)= \sum_{l:\xv\in X_l} \pi_l\prod_{j=1}^p \hat{p}_{l,j}(x_j),
    \label{eq:FORDE}
    \end{equation}
where $X_l$ is the hyperrectangle defined by the $l$-th leaf, the corresponding mixture weights $\pi_l$ are calculated as the share of real datapoints that fall into leaf $l$ normalized over all trees, and $\hat{p}_{l,j}$ are (locally) estimated univariate density/mass functions for the $j$-th feature in leaf $l$. The convergence of FORDE to the real data distribution of $\Xv$ for infinite data is proven under some mild conditions in Watson et al. \citep{watson2023}. A conditional density under a set of conditions $\mathcal{C}$, e.g., fixed values or intervals for certain features $C \subseteq \{1, ..., p\}$, can be derived from \Cref{eq:FORDE} in the following way:
    \begin{equation}
        \text{FORDE}(\xv \mid \mathcal{C})\coloneqq\hat{p}(\xv \mid \mathcal{C})= \sum_{l:\xv\in X_l} \pi'_l \prod_{j=1}^p \hat{p}_{l,j}(x_j \mid \mathcal{C}_j),
    \label{eq:cFORDE}
    \end{equation}
where $\mathcal{C}_j \subseteq \mathcal{C}$ denotes the subset of conditions concerning feature $j \in C$, and the mixture weights $\pi'_l$ are updated to reflect how likely their corresponding leaves fulfill the condition. 
More formally, the mixture weights
are updated and normalized using the univariate marginals by 
\begin{equation*}
\pi'_l \coloneqq \frac{\pi_l \prod_{j=1}^p \hat{p}_{l,j}(\mathcal{C}_j) }{ \sum_{m:\xv\in X_m} \pi_m \prod_{j=1}^p \hat{p}_{m,j}(\mathcal{C}_j)}
\end{equation*}
if the denominator does not equal $0$ and by $\pi'_l \coloneqq 0$ otherwise. %
Note that in the case of conditioning on a fixed value or interval for a continuous feature $j$, the univariate densities $\hat{p}_{l,j}$ collapse to the indicator function $\mathbbm{1}_{\mathcal{C}_j}$ or the unconditional densities truncated on the conditioning interval, respectively.
    
\paragraph{Generative modeling with FORGE} 
is based on drawing a leaf $l$ from the forest according to the mixture weights in FORDE and sampling feature values from the estimated univariate (conditional) densities $\hat{p}_{l,j}$. Thereby, FORGE allows to draw samples that adhere to FORDE as an approximation to the real distribution of $\Xv$ or $\Xv \mid \mathcal{C}$.

\section{Methods}
\label{sec:methods}
Our proposal is to leverage ARF for the efficient generation of counterfactual explanations, i.e., countARFactuals, in mixed tabular data settings. More specifically, we use and modify ARF to account for the desiderata that we discussed in \Cref{subsec:counterfac}:
\begin{enumerate}
    \item \textbf{Validity:} We train ARF on $\D$ but replace the target $Y$ with the predictions $\hat{Y}$. Here, $\hat{Y}$ is treated just as any other feature in the data. Since FORGE allows for conditional sampling, we can sample from $\Xv$ conditioned on our desired outcomes $\hat{Y}\in Y_{des}$. Note, however, that ARF may not learn a perfect representation of the prediction function $\hat{Y}:=\hat{f}(\Xv)$. It therefore is not guaranteed that ARF-samples are valid, it only becomes more likely. In our algorithms, we only return those candidates with predictions in $Y_{des}$.
    \item \textbf{Proximity:} We restrict the output of our two methods to those counterfactuals in the Pareto set, defined over the four objectives of \Cref{subsec:counterfac}, including proximity (\Cref{eq:moc-prox}). In the first algorithm described below, we additionally use ARF combined with MOC, which accounts for proximity, as described in \Cref{subsec:counterfac}.
    \item \textbf{Plausibility:} ARF allows us to both evaluate the plausibility of data points using FORDE (which is also used to determine the returned Pareto set) and efficiently generate plausible data with FORGE.
    \item \textbf{Sparsity:} FORGE allows to sample feature values $X_S$ based on the observation $X_C=x_C$. By fixing certain features $C$ to the value of $\xv^*_C$, we only change feature values in the sparse set $S:=\lbrace 1,\dots,p\rbrace\setminus C$.
\end{enumerate}
With the desiderata in place, several decisions need to be made: Should we integrate plausibility via density estimation (FORDE) or generative modeling (FORGE)? What is an optimal trade-off between proximity and other objectives, such as plausibility and sparsity? How should we search for the conditioning set $C$ for features that should not be changed? In the following, we provide two algorithms that decide on these questions in different ways. The first integrates ARF into MOC (\Cref{subsec:algo1-MOC+ARF}). The second uses ARF as a standalone counterfactual generator (\Cref{subsec:algo2-onlyARF}).

\subsection{Algorithm 1: Integrating ARF into MOC}
\label{subsec:algo1-MOC+ARF}
In MOC's optimization problem (\Cref{eq:mutli-objective}), we substitute the plausibility measure (\Cref{eq:moc-plaus}) by the density estimator of FORDE (\Cref{eq:FORDE}). Since the individual objectives in MOC must map to a zero-one interval (with low values denoting desired properties), we transform $\hat{p}(\xv)$, as estimated by FORDE, with the negative exponential function 
\begin{equation}
    o_{\text{plaus}}^*(\xv):=e^{-\hat{p}(\xv)}.
    \label{eq:arf-plaus}
\end{equation}
We use FORGE as described above to sample plausible candidates in MOC in the mutation step of the NSGA-II. This is a strategy to efficiently limit the search space of MOC to plausible counterfactuals. Concerning sparsity, we find the conditioning set $C$ through iterated mutation and recombination, just like in MOC, and we select candidates using NSGA-II  according to optimality and diversity. 
At last, the output comprises only the valid Pareto-set of counterfactuals over the generations, i.e., counterfactuals that have a prediction in $Y_{des}$ and are not dominated by other candidates that were generated.
For details, we refer to the pseudocode in \Cref{ap:algo1-MOC+ARF}. 

\subsection{Algorithm 2: ARF is all you need}
\label{subsec:algo2-onlyARF}
For this algorithm, we leverage the ability of our modified ARF sampler to directly and efficiently generate many relevant counterfactuals. As described above, the modified FORGE method allows to generate plausible data points. To enforce sparsity, we sample $m$ features with probabilities according to their local feature importance, calculated as the standard deviation of the individual conditional expectation (ICE) curve \citep{dandl2020multi,goldstein2015}.
The $m$ selected features describe the features $S$ we aim to change because they, according to the local feature importance, impact the prediction the most. The remaining features then form the conditioning set $C=\lbrace 1,\dots,p\rbrace\setminus S$.

As for Algorithm 1, we output only the valid and Pareto-optimal set of counterfactuals. 
The pseudocode for this method is given in \Cref{ap:algo2-onlyARF}.

\section{Experiments}
\label{sec:experiments}

We evaluate the quality of our proposed methods with respect to the following research questions:
\begin{enumerate}
\setlength{\itemindent}{1.8em}
\item[RQ (1)] Do our proposed ARF-based methods generate more plausible counterfactuals compared to competing methods without major sacrifices in sparsity ($o_{\text{sparse}}$), proximity ($o_{\text{prox}}$) and the runtime?
\item[RQ (2)] Does $o^*_{\text{plaus}}$ (\Cref{eq:arf-plaus}) better reflect the true plausibility compared to $o_{\text{plaus}}$ (\Cref{eq:moc-plaus})?
\end{enumerate}
To objectively evaluate the plausibility of the generated counterfactuals, we require access to the ground-truth likelihood. Because ground-truth likelihoods are usually unavailable for real-world data, we evaluate our methods on synthetic data. An illustrative real-world application follows in \Cref{sec:real-data-example}.

\subsection{Data-generating process}
For the experiments, we constructed three illustrative two-dimensional datasets, namely \texttt{cassini} (inspired by \cite{mlbench2021}), \texttt{two sines} (inspired by the two moons dataset), and \texttt{three blobs} (inspired by \cite{pawelczyk2020learning}). Moreover, we generated four da\-ta\-sets from randomly sampled Bayesian networks of dimensionality $5$, $10$, and $20$, namely  \texttt{bn\_5}, \texttt{bn\_10}, and \texttt{bn\_20}, which all include both continuous and categorical features as well as nonlinear relationships. An XGBoost model was fitted on sampled da\-ta\-sets $D_{\text{train}}$ of size $5\,000$ \citep{Chen2016_xgboost}. For each data-generating process (DGP), ten additional points were sampled as instances of interest $\xv^*$. The counterfactual generation methods received access to newly sampled datasets $\D$ of size $5\,000$. Details on the dataset generation and model fit can be found in Appendix \ref{appendix:synthetic-data} and in the repository accompanying this paper.\footnote{\url{https://github.com/bips-hb/countARFactuals}.}

\subsection{Competing methods}
We compare our proposed MOC version based on ARF of \Cref{subsec:algo1-MOC+ARF} (referred to as MOCARF) and the standalone ARF generator of \Cref{subsec:algo2-onlyARF} (referred to as ARF) to the following competitors: MOC and MOCCTREE (MOC with a conditional sampler, see \Cref{subsec:counterfac}) \cite{dandl2020multi} and NICE \cite{brughmans2023nice} with a plausibility reward function (see Equation~(4) in \cite{brughmans2023nice}). 
NICE generates counterfactuals by iteratively replacing one feature after the other in $\xv^*$ by the values of $\xv^\text{nn}$, which denotes a nearest neighbor of $\xv^*$ in $\D$ with $\fh(\xv^\text{nn}) \in Y_{des}$. In each iteration, the algorithm keeps the feature change with the highest plausibility reward. 

To allow for a fair comparison, all methods generate a \textit{set} of counterfactual candidates. For NICE, we apply the extension of Dandl et al. \citep{dandl2023counterfactuals}; instead of stopping the search once the point with the highest reward has a prediction in $Y_{des}$, the search continues until $\xv^\text{nn}$ is recovered and all intermediate instances with predictions in $Y_{des}$ are returned. 
If possible, we selected the hyperparameters for the methods such that each method generated an equal number of candidates -- namely, $1\,000$.\footnote{Specifying the exact number was possible for all methods besides NICE \citep{dandl2023counterfactuals}.}
ARF requires a maximum set size for $S$, reflecting how many features are maximally allowed to be changed. We set it according to the number of features $p$ as $m_{max} := min(\lceil \sqrt{p} + 3 \rceil, p)$. 
Since also for all MOC-based methods the maximum number can be specified, we used the same $m_{max}$ for MOC, MOCARF and MOCCTREE.
For the evaluation, we focused only on the unique counterfactuals that have predictions in $Y_{des}$. We further reduced this set to the Pareto set, i.e., the set of counterfactuals that are nondominated according to proximity ($o_{\text{prox}}$), sparsity ($o_{\text{sparse}}$) and plausibility. The definition of the plausibility objective differed between the methods, with $o^\star_{\text{plaus}}$ for ARF and MOCARF, $o^{\text{plaus}}$ for MOC and MOCCTREE, and the autoencoder reconstruction error for NICE (as proposed by \cite{brughmans2023nice}).

\subsection{Evaluation criteria}
To answer RQ (1), we evaluated the generated counterfactuals with respect to the ground-truth likelihood (denoted as plausibility, in the following), validity $o_{\text{valid}}$, proximity $o_{\text{prox}}$ and sparsity $o_{\text{sparse}}$ (see \Cref{subsec:counterfac}). We aggregated the results per method, dataset and instance of interest $\xv^\star$ by computing (scaled) dominated hypervolumes \citep{zitzler1998}. We also measured the number of nondominated counterfactuals and the runtime.
To investigate the trade-off between plausibility and proximity, we also computed median attainment surfaces according to L{\'o}pez-Ib{\'a}{\~{n}}ez et al. \citep{lopez2010} for each method and dataset. It reveals how the two objectives are distributed on average over the different $\xv^*$.
To answer RQ (2), all generated counterfactuals were evaluated with respect to $o_{\text{plaus}}^\star$ and $o_{\text{plaus}}$. Per method, dataset and $\xv^\star$, we computed Spearman-rank correlations between the true plausibility and $o_{\text{plaus}}^\star$ and between the true plausibility and $o_{\text{plaus}}$. 
With a Wilcoxon signed rank test, we tested whether $o_{\text{plaus}}^*$ has higher correlations to the true plausibility than $o_{\text{plaus}}$.

\subsection{Results}
\begin{figure}[ht]
    \centering
    \includegraphics[width = 1\textwidth]{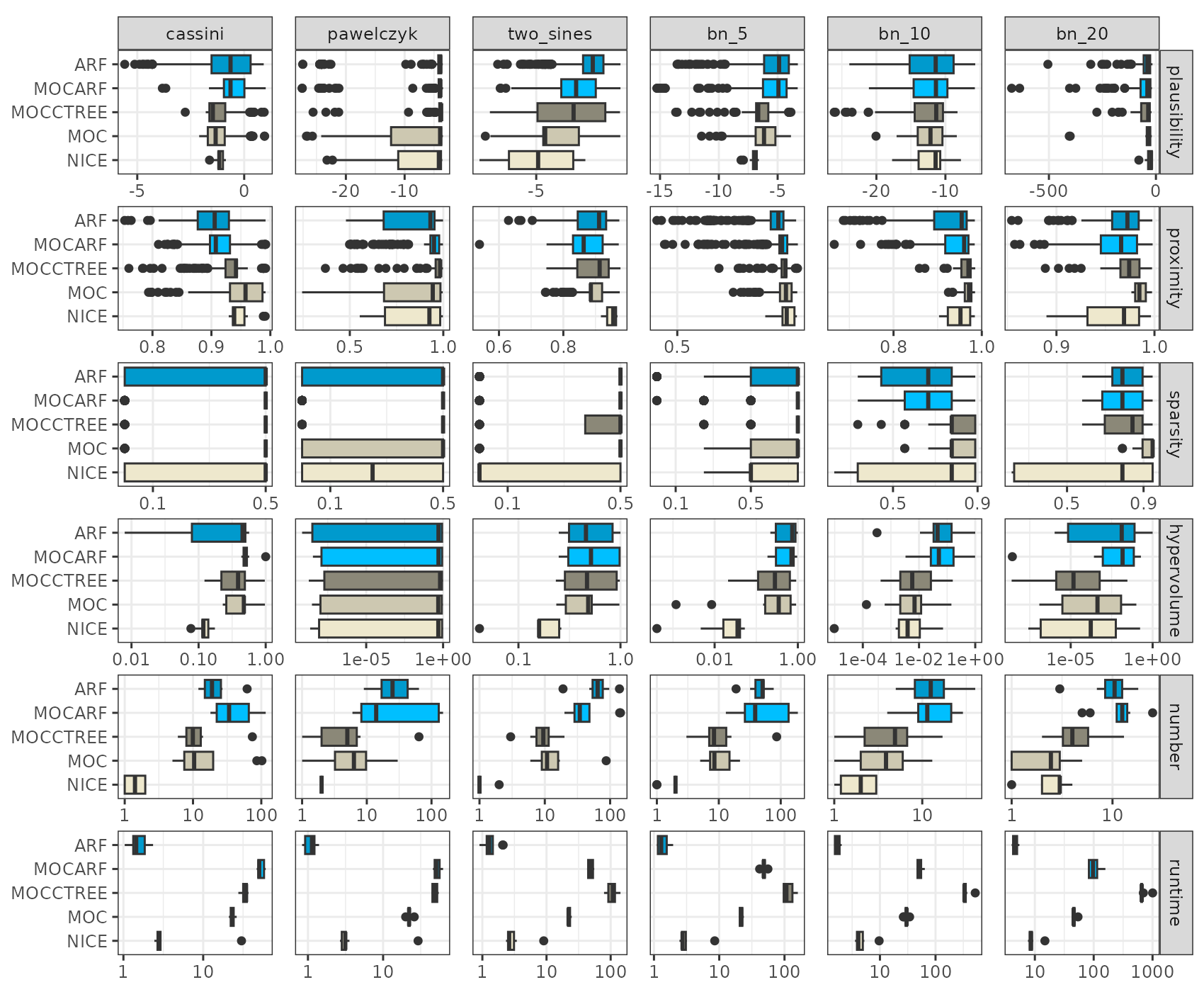}
    \caption{Boxplots of the plausibility, proximity ($1 - o_{\text{prox}}$), sparsity ($1 - o_{\text{sparse}}$), hypervolume, number of counterfactuals and runtime for each method and dataset. Higher values are better, except for runtime.}
    \label{fig:obj-hv}
\end{figure}

\Cref{fig:obj-hv} presents the results for RQ (1) and shows the objective values per counterfactuals as well as the hypervolume, number of counterfactuals and runtime.
On average, ARF and MOCARF generated more plausible counterfactuals compared to the other MOC-based approaches and NICE. 
In alignment with previous literature \citep{dandl2020multi,del_ser_2024}, our results suggest that higher plausibility might be associated with lower proximity and sparsity.
For further investigations on the trade-offs, \Cref{fig:eaf} and \Cref{fig:eaf-spars} in \Cref{ap:extraresults} detail the median attainment surfaces per dataset and method. 
The plots reveal that ARF and MOCARF on average dominate the other methods in proximity, sparsity and plausibility, with the differences being greatest in plausibility.
The hypervolume was on average similar for the different methods for low-dimensional datasets (ARF had lower hypervolumes in \texttt{cassini} due to its inferiority in proximity and sparsity), for higher-dimensional datasets (\texttt{bn\_10} and \texttt{bn\_20}), ARF and MOCARF performed better than the competing methods. 
Concerning runtime, ARF generated counterfactuals the fastest on average, followed by NICE and MOC. 
MOCARF was faster than MOCCTREE for datasets with more than two features.
The runtime differences increased with higher dimensional data. 
On average, ARF and MOCARF generated the largest set of nondominated counterfactuals compared to the other methods.

Considering RQ (2), the Wilcoxon rank sum test had a p-value close to $0$ ($7.16e-06$), i.e., the correlation of our proposed plausibility measure $o_{\text{plaus}}^*$ to the true plausibility was significantly higher than that of $o_{\text{plaus}}$. The median correlation to the true plausibility over all methods and datasets was 0.84 for $o_{\text{plaus}}^*$ and 0.69 for $o_{\text{plaus}}$.

Overall, our study shows that on average our proposed methods -- ARF and MOCARF -- generate a more plausible set of counterfactuals compared to our competitors without major sacrifices in sparsity and proximity. Notably, ARF achieves this with superiority in runtimes.

\section{Real Data Example}
\label{sec:real-data-example}
We illustrate our approach on the publicly available coffee quality dataset\footnote{\url{https://github.com/jldbc/coffee-quality-database}}. The data details the characteristics of several Arabica coffee beans, such as the country of origin and altitude at which the beans were cultivated. Further, the dataset includes information on a quality review score (\textit{cup points}) specified by an expert jury within the Coffee Quality Institute \citep{coffee_institute}.

In this example, we use a random forest to predict coffee quality from selected, actionable characteristics of the coffee beans. For simplicity, we binarize the target score \textit{cup points}. Aiming for balanced classes of \textit{good} and \textit{bad} quality, we use the dataset’s median value of  \textit{cup points} as a cut-off point, i.e., 
\begin{equation}
  \texttt{quality} =
    \begin{cases}
      \text{\textit{good}} & \text{if \textit{cup points} $\geq$ median(\textit{cup points})}\\
      \text{\textit{bad}} & \text{otherwise}.
    \end{cases}       
\end{equation}

For illustration, we generate counterfactual explanations for an instance of bad coffee quality,  answering the question: Which characteristics would need to be changed to rate as good quality coffee? 

This example illustrates the importance of taking into account the multiple objectives of counterfactual explanations, such as sparsity and plausibility. For example, a company that aims to improve the quality of their coffee may want to make as sparse changes to the coffee characteristics as possible for economic reasons. Similarly, some changes might not be plausible, think of changing the country of origin independently of the altitude of the coffee plantations or the variety of beans cultivated in the respective country (since the variety must suit the natural conditions in the respective country).

\begin{figure}[htbp]
    \centering
    \includegraphics[clip, trim=3.8cm 33cm 4cm 4cm, width=1.00\textwidth]{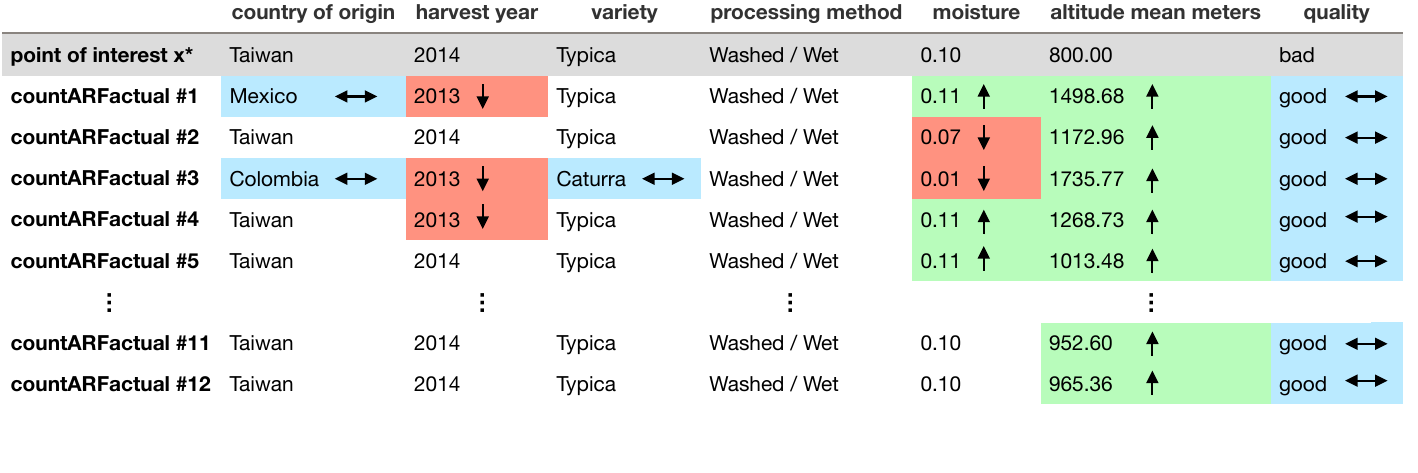}
    \caption{Exemplary countARFactuals for an instance of \textit{bad} coffee quality. Arrows indicate changes in comparison to $\xv^*$, i.e., a feature's value increase $\uparrow$, decrease $\downarrow$ or change in category $\leftrightarrow$. }
    \label{fig:coffee}
\end{figure}

The generation of counterfactual explanations in this example is performed using Alg. 2  detailed in Section \ref{subsec:algo2-onlyARF}.  In Figure \ref{fig:coffee}, we present a set of the generated countARFactuals explanations for an instance of coffee beans belonging to the \textit{bad} class that originate from Taiwan. 
Figure~\ref{fig:coffee} illustrates that countARFactuals yield plausible counterfactual explanations. For instance, for countARFactual \#3, both the country of origin is changed from Taiwan to Colombia and the variety from Typica to Caturra. This seems reasonable because Typica was grown in only few Colombian instances in the training dataset, and instead, Caturra was the most frequently grown variety in Colombia. Further, the altitude at which the beans are grown is elevated only a little within Taiwanese countARFactuals (\# 4 - 12), but more drastically for countries that -- given the data -- grow coffee on higher altitudes on average, such as Mexico (\# 1) and Colombia (\# 3).

\section{Discussion}
\label{sec:conclusion}
In this paper we show that adversarial random forests (ARF) can be modified to generate plausible counterfactuals, both as a subroutine to multi-objective counterfactual explanations (MOC) and as a standalone approach. Our experiments in \Cref{sec:experiments} demonstrate that ARF can improve the plausibility of counterfactuals and the efficiency in their generation without substantially sacrificing other desiderata such as proximity and sparsity.
In contrast to other generative modeling approaches for plausible counterfactuals, ARF handles mixed tabular data directly without, e.g., one-hot-encoding categorical features, thereby improving data-efficiency. Moreover, ARF-based counterfactual generation allows for sparsity via conditional sampling and is an off-the-shelf methodology that requires minimal efforts in tuning and computational resources.

Our work faces some limitations. For example, we define the plausibility of counterfactuals via the joint density. However, as highlighted by Keane et al. \cite{keane2021if}, there are different conceptualizations of plausibility, for example, based on the feasibility of actions or user perceived plausibility \citep{karimi2022survey,konig2023improvement,poyiadzi2020face}. One might even question if staying in the manifold is always desirable, e.g., if changing the class requires extrapolation so should our counterfactuals.
It should be noted that plausible counterfactuals, in general, cannot be interpreted as action recommendations. Although they provide hints about which alternative feature values would yield acceptance by the predictor, they do not guide the user on which interventions yield the desired change in the real world. To guide action, causal knowledge is required \cite{karimi2021algorithmic}. Furthermore, in the context of recourse, \textit{improvement} of the underlying target is more desirable than \textit{acceptance} by a specific predictor, which counterfactual explanations do not target \cite{konig2023improvement}.

Proximity and plausibility are conflicting objectives \citep{brughmans2023nice,dandl2020multi}. Oftentimes, there is only little data close to the decision boundary, and jumping just over the boundary can lead to implausible counterfactuals \citep{freiesleben2022intriguing}. A trade-off between the two objectives is desirable, which we implicitly address by generating a Pareto-optimal set of diverse counterfactuals. In future work, one could already incorporate such trade-offs in the counterfactual generation, e.g., by a parameter that
directly controls for the proximity-plausibility trade-off. 
One option would be to set a threshold for plausibility instead of a trade-off parameter, as suggested by Brughmans et al. \cite{brughmans2023nice}.

Like all works on counterfactual explanations, we face the Rashomon effect: There exist many plausible counterfactuals that explain the same data point. This raises the question of which one we should show to the user \citep{keane2021if,verma2020counterfactual}? As a bottom line, we return only the Pareto-optimal set of counterfactuals, which at least guarantees that no strictly dominated option is shown. In future work, integrating user preferences or considering additional objectives may improve the final selection.

Our framework is tailored for mixed tabular data settings. For other data modalities like image or text data, we advise for using neural network based approaches for density estimation and generative modeling such as VAEs and GANs. 
Finally, our framework is designed for binary classification and regression but can be extended to multi-class classification.

In future work, we plan to investigate the role of the ML model in the ARF approach to counterfactuals. We could also generate counterfactuals with ARF without the model by directly training ARF on $Y$ rather than the predictions $\hat{Y}$. We would then get plausible counterfactuals that hint towards improvement instead of acceptance \citep{konig2023improvement}. While such counterfactuals appear different from those discussed in the XAI literature so far, in fact, they essentially just turn the generative model that conditions on $\Xv=\xv$ into a prediction algorithm.

\section*{Acknowledgments}
MNW and KB were supported by the German Research Foundation (DFG), Grant Number 437611051. MNW was  supported by the German Research Foundation (DFG), Grant Number 459360854. KB was supported by a PhD grant of the Minds, Media, Machines Integrated Graduate School Bremen. MNW and JK were supported by the U Bremen Research Alliance/AI Center for Health Care, financially supported by the Federal State of Bremen. GK and TF were supported by the German Research Foundation through the Cluster of Excellence “Machine Learning - New Perspectives for Science" (EXC 2064/1 number 390727645). TF has been supported by the Carl Zeiss Foundation through the project “Certification and Foundations of Safe Machine Learning Systems in Healthcare”. 
\bibliographystyle{unsrt}  
\bibliography{main.bib}  

\newpage
\appendix

\section{Algorithm 1: Integrating ARF into MOC}
\label{ap:algo1-MOC+ARF}

The following pseudocode is based on Algorithm~1 in \citep{dandl2023counterfactuals}.
Blue lines highlight the steps that differ from the original MOC algorithm proposed by \citep{dandl2020multi}.

\begin{algorithm}[ht!] \label{alg::moc_arf}
 		\caption{MOC with ARF-based Sampler and Evaluation}
 		\textbf{Inputs:} \\
 		Datapoint to explain prediction for $\mathbf{x}^\star \in \mathcal{X}$ \\
		Desired outcome (range) $Y_{des}$ \\
 		Prediction function $\hat{f}:\Xspace \rightarrow \R$ \\
 		Observed data $\D$ \\
        \textcolor{blue}{ARF $\hat{g}^*$ trained on $(\xi, \fh(\xi))_{i = 1}^n$ with $\xi \in \D$} \\
 		Number of generations $n_{\text{generations}}$ \\
 		Size of population $\mu$ \\
 		Recombination and mutation methods including probabilities \\
 		Selection method for features in the conditioning set and initialization method \\
        Stopping criterion \\
 		(Additional user inputs, e.g., range of numerical features, immutable features, distance function, see \citep{dandl2020multi})\\
 		\begin{algorithmic} 
 			\item[1:]Initialize population $P_0$ with $|P_0| = \mu$ (ICE-curve-based, see \cite{dandl2020multi})
 			\item[2:] Evaluate candidates according to four objectives:
                \begin{itemize}
                    \itemsep-.1em
                    \item Validity ($L_1$)
                    \item Sparsity ($L_0$)
                    \item Proximity (Gower distance)
                    \item Plausibility \textcolor{blue}{(ARF-based likelihood transformed with $e^{-x}$)}
                \end{itemize}
 			\item[3:] Set $t = 0$
 			\item[4:] \textbf{for} $r \in \{1, ..., n_{\text{iterations}}\}$
 			\item[5:] \hspace{1cm} $C_{t} =$ \texttt{create\_offspring}($P_{t}$), $|C_t| = \mu$ with given probabilities 
                \begin{enumerate}[leftmargin=2cm]
                \itemsep-.1em
                    \item Select best candidates (acc. to validity objective)
                    \item Recombine these pairwise 
                    \item \textcolor{blue}{Mutate values jointly using $\hat{g}^*$: generate new datapoints with FORGE}
                \end{enumerate}
 			\item[6:] \hspace{1cm} Combine parents and offspring $R_t = C_t \cup P_t$
 			\item[7:] \hspace{1cm} Assign candidates to a front according to their objective values: \\
 			\hspace{1.1cm} $(F_1, F_2, ..., F_m) = $ \texttt{ nondominated\_sorting($R_t$)}
 			\item[8:] \hspace{1cm} \textbf{for} $i = 1, ..., m$
 			\item[9:] \hspace{2cm} Sort candidates acc. to diversity (objective and feature space): \\
 			 \hspace{2.1cm} $\tilde{F}_i$ = \texttt{crowding\_distance\_sort($F_i$)}
            \item[10:] \hspace{.9cm} \textbf{end for}
 			\item[11:] \hspace{.9cm} Set $P_{t+1} = \emptyset$ and $i = 1$
 			\item[12:] \hspace{.9cm} \textbf{while} $|P_{t+1}| + |\tilde{F}_i| \le \mu$ 
 			\item[13:] \hspace{2cm} $P_{t+1} = P_{t+1} \cup \tilde{F}_i$
 			\item[14:] \hspace{2cm} i = i + 1
			\item[15:] \hspace{.9cm} \textbf{end while}
			\item[16:] \hspace{.9cm} Choose first $\mu-|P_{t+1}|$ elements of $\tilde{F}_i$: $P_{t+1} = P_{t+1} \cup \tilde{F}_i[1:(\mu-|P_{t+1}|)]$
 			\item[17:] \hspace{.9cm} $t = t + 1$
 		    \item[18:] \textbf{end for}
 		    \item[19:] Return unique, non-dominated candidates of $\bigcup_{k = 0}^{t} P_{k} \setminus \mathbf{x}^\star$ with $\fh(\xv_{CF}) \in Y_{des}$
	    \end{algorithmic}
\end{algorithm}  

\newpage
\section{Algorithm 2: ARF is all you need}
\label{ap:algo2-onlyARF}

\begin{algorithm}[ht!] \label{alg::only_arf}
 		\caption{ARF-based Counterfactual Generator}
 		\textbf{Inputs:} \\
 		Datapoint to explain prediction for $\mathbf{x}^\star \in \mathcal{X}$ \\
		Desired outcome (range) $Y_{des}$ \\
 		Prediction function $\hat{f}:\Xspace \rightarrow \R$ \\
 		Observed data $\D$ \\
        ARF $\hat{g}^*$ trained on data $(\xi, \fh(\xi))_{i = 1}^n$ with $\xi \in \D$ \\
        Maximum number of feature changes $m_{max}$ \\
 		Number of iterations $n_{\text{iterations}}$ \\
 		Number of samples generated in each iteration $n_{synth}$\\
 		(Additional user inputs, e.g., immutable features) \\
 		\begin{algorithmic} 
 			\item[1:] Derive local importances $(\text{fi}_j)_{j = 1}^p$ for each feature $j \in \{1, ..., p\}$ (ICE-curve-based, see \cite{dandl2020multi})
            \item[2:] \textbf{for} $r \in \{1, ..., n_{\text{iterations}} \}$
            \item[3:] \hspace{1cm} $m \leftarrow \texttt{sample}(1, ..., m_{max})$
 			\item[4:] \hspace{1cm} Select set $C \subset \{1, ..., p\}$ by randomly sampling $m$ features with probability \\
    \hspace{1.1cm} proportional to how \textit{unimportant} feature is
 			\item[5:] \hspace{1cm} $CF \leftarrow$ sample $n_{synth}$ observations with FORGE derived from $\hat{g}^*$ under \\
            \hspace{1.1cm} condition that $\forall j \in C: X_j = x_j^\star$ \& $\hat{Y} \in Y_{des}$ 
            \item[6:] $\Xv_{CF} \leftarrow (\Xv_{CF}, CF)$
 		    \item[7:] \textbf{end for}
 		    \item[8:] Return unique, nondominated candidates $\xv_{CF} \in \Xv_{CF}$ with $\fh(\xv_{CF}) \in Y_{des}$
	    \end{algorithmic}
 \end{algorithm}

 \newpage
 \section{Synthetic Data}
 \label{appendix:synthetic-data}

 As follows, we describe the three illustrative datasets as well as the sampling of the randomly generated data-generating processes. The code that was used to generate the datasets and pair plots visualizing their distribution can be found in the repository accompanying the paper (\url{https://github.com/bips-hb/countARFactuals}).\footnote{For an explanation of how to run the code, we refer to \texttt{python/README.md}. The visualization can be found in the folder \texttt{python/visualizations/}.}

 \subsection{Illustrative datasets}

\paragraph{Cassini}{ The DGP, inspired by \cite{mlbench2021}, is defined as follows:
\begin{align*}
    Y &\sim Y_1 + Y_1 Y_2 \quad \text{with} \quad Y_1 \sim Bern(2/3),\; Y_2 \sim Bern(0.5)\\
    X_1|Y_1 &\sim 
    \begin{cases}
        N(0, 0.2), &\text{if } Y_1 = 0\\
        N(0, 0.5), &\text{otherwise}
    \end{cases}\\
    X_2 | X_1, Y_1, Y_2 &\sim 
    \begin{cases}
        N(0, 0.2) &\text{if } Y_1 = 0\\
        N((-1)^{Y_2}cos(X_1), 0.2) &\text{otherwise}
    \end{cases}
\end{align*}
}
\paragraph{Two Sines}{ The DGP, inspired by the two moons dataset, is specified as:
\begin{align*}
    Y \sim Bern(0.5), \quad
    X_1 | Y \sim N(Y, 3.0), \quad
    X_2|Y, X_1 \sim N(sin(X_1) - 2Y +1, 0.3)
\end{align*}
}
\paragraph{Pawelczyk}{The DGP, taken from \cite{pawelczyk2020learning}, is defined below, where $I_2$ refers to the $2x2$ identity matrix and $Cat({\textstyle\frac{1}{3}, \frac{1}{3}, \frac{1}{3}})$ to the uniform categorical distribution with values $0, 1, 2$.
\begin{align*}
    L &\sim Cat({\textstyle\frac{1}{3}, \frac{1}{3}, \frac{1}{3}})\\
    X | \mu &\sim N(\mu, I_2 ), \quad \mu | L =
    \begin{cases}
        (-10, 5)^T & \text{if } L=0\\
        (0,5)^T & \text{if } L=1\\
        (0,0)^T & \text{otherwise}\\
    \end{cases}\\
    Y (X) &:= X_2 > 6
\end{align*}
}

\subsection{Randomly generated DGPs}

For the generation of \texttt{bn\_5}, \texttt{bn\_10}, and \texttt{bn\_20}, we randomly sample Bayesian networks with categorical and continuous distributions as well as linear and nonlinear relationships.
\begin{enumerate}
    \item First, we randomly sample a Directed Acyclic Graph (DAG) using the \texttt{networkx} package. We select $Y$ as the root node. To make sure that $Y$ is related to many of the features, for each node that is not directly neighboring $Y$, a directed edge is added with probability $0.5$ (directed such that the graph remains acyclic).
    \item From all nodes, $20\%$ are randomly selected to be categorical nodes; $Y$ is always selected to be a categorical node.
    \item For every node $j$, an aggregation function $g$ is sampled that maps the parent values $x_{pa(j)}$ to an aggregate, which then parameterizes the distribution of the respective node.
    \begin{equation*}
        g(x) = \beta + \beta_1h(x) + \beta_2h(x)^2 \quad \text{with} \quad h(x) = sin\left(\sum_{i \in pa(j)} w_i x_i\right)
    \end{equation*}
    The weights $w$ are sampled from $Unif(-1, 1) + 3Bern(3/d)$. To make it more likely that $Y$ can be predicted well from its covariates, weights concerning $Y$ are increased by $d \sim Unif(3,4)$ with probability $0.1$. The weight vector $w$ is normalized. The coefficients $\beta$ are sampled from $Unif(-1,1)$.
    \item If the feature is categorical, the respective Bernoulli is parameterized with the sigmoid of the aggregate of the parents $Bern(\varsigma(g(x)))$. Continuous features follow $N(\mu, \sigma)$ with $\mu \sim N(g(x), 1)$ and $\sigma \sim N(0,2)$.
\end{enumerate}
To ensure that a prediction model fitted on the data can discriminate between the classes and that changing the prediction to the desirable class is feasible, we randomly generated datasets until we found one with balanced labels ($0.4 < E[Y] < 0.6$), and where a \texttt{xgboost} model demonstrated good accuracy ($> 0.95$) and balanced predictions ($0.3 < E[\hat{Y}] < 0.7$).

\newpage
\section{Additional empirical results}
\label{ap:extraresults}

\begin{figure}[!h]
\vspace{-7mm}
    \centering %
\begin{subfigure}{0.3\textwidth}
  \includegraphics[width=\linewidth]{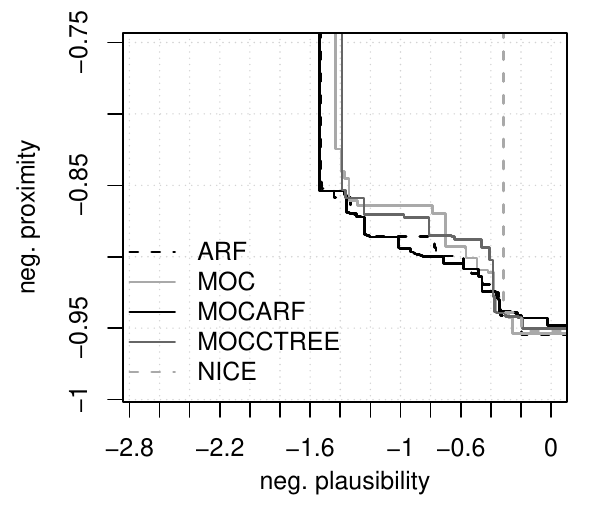}
  \vspace{-7mm}
  \caption{\texttt{cassini}}
  \label{fig:11}
\end{subfigure}\hfil %
\begin{subfigure}{0.3\textwidth}
  \includegraphics[width=\linewidth]{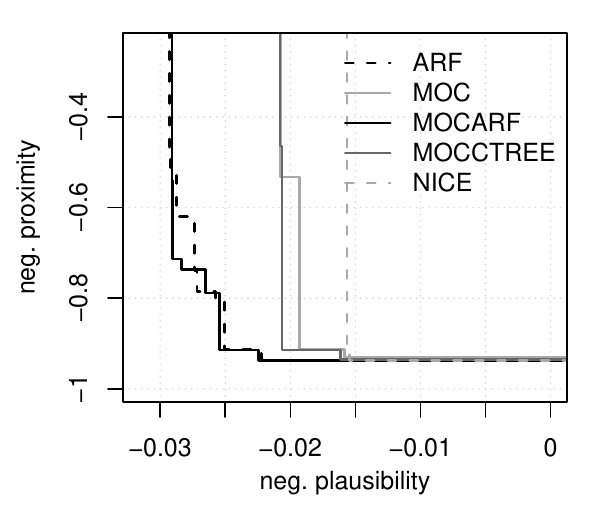}
  \vspace{-7mm}
  \caption{\texttt{pawelczyk}}
  \label{fig:22}
\end{subfigure}\hfil %
\begin{subfigure}{0.3\textwidth}
  \includegraphics[width=\linewidth]{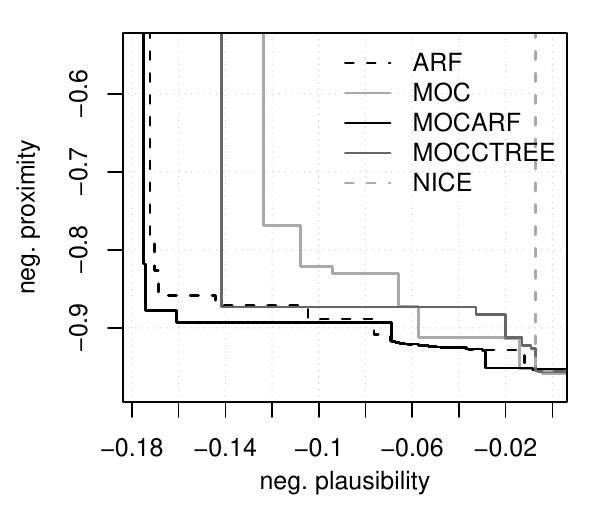}
  \vspace{-7mm}
  \caption{\texttt{two\_sines}}
  \label{fig:33}
\end{subfigure}

\medskip
\begin{subfigure}{0.3\textwidth}
  \includegraphics[width=\linewidth]{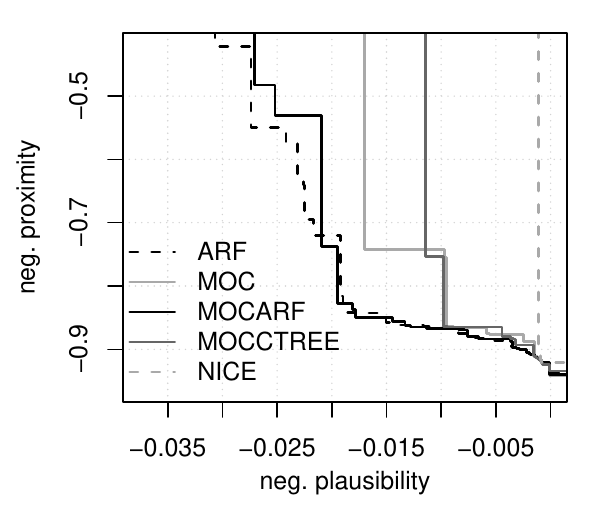}
  \vspace{-7mm}
  \caption{\texttt{bn\_5}}
  \label{fig:44}
\end{subfigure}\hfil %
\begin{subfigure}{0.3\textwidth}
  \includegraphics[width=\linewidth]{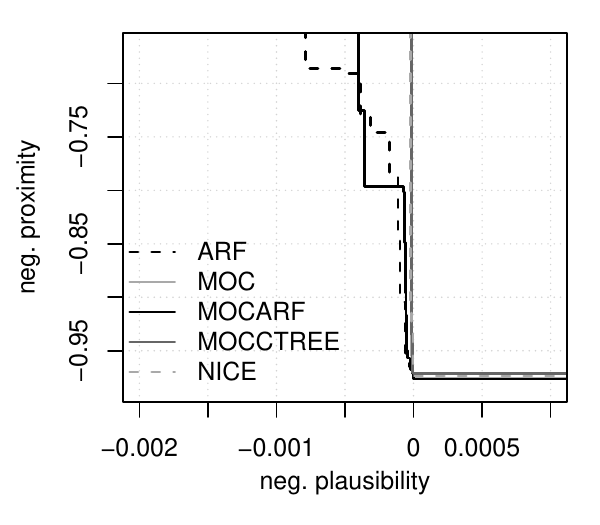}
  \vspace{-7mm}
  \caption{\texttt{bn\_10}}
  \label{fig:55}
\end{subfigure}\hfil %
\begin{subfigure}{0.3\textwidth}
  \includegraphics[width=\linewidth]{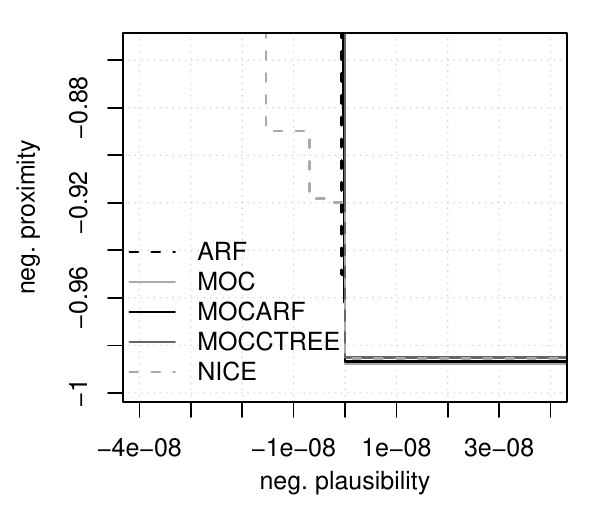}
  \vspace{-7mm}
  \caption{\texttt{bn\_20}}
  \label{fig:66}
\end{subfigure}
\caption{Median empirical attainment function \citep{lopez2010} for the negative plausibility and negative proximity. Lower values are better.}
\label{fig:eaf}
\end{figure}

\begin{figure}[!h]
\vspace{-7mm}
    \centering %
\begin{subfigure}{0.3\textwidth}
  \includegraphics[width=\linewidth]{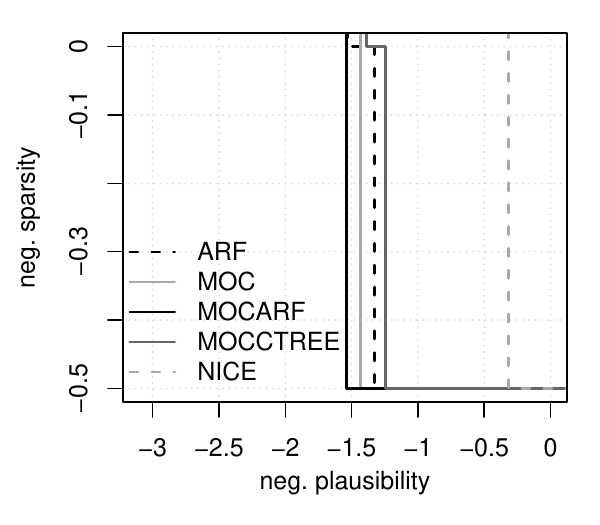}
  \vspace{-7mm}
  \caption{\texttt{cassini}}
  \label{fig:1}
\end{subfigure}\hfil %
\begin{subfigure}{0.3\textwidth}
  \includegraphics[width=\linewidth]{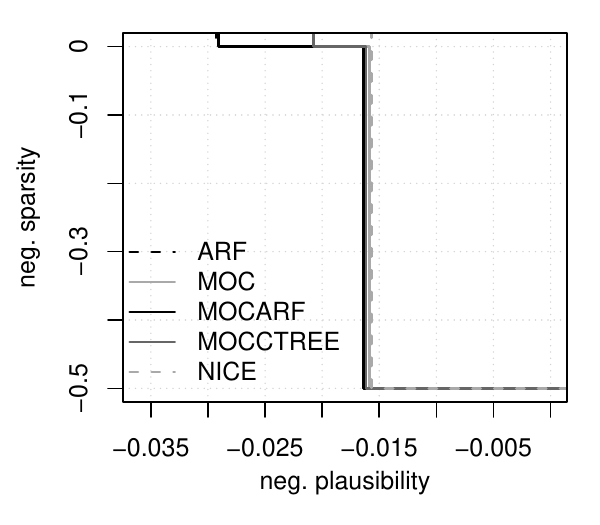}
  \vspace{-7mm}
  \caption{\texttt{pawelczyk}}
  \label{fig:2}
\end{subfigure}\hfil %
\begin{subfigure}{0.3\textwidth}
  \includegraphics[width=\linewidth]{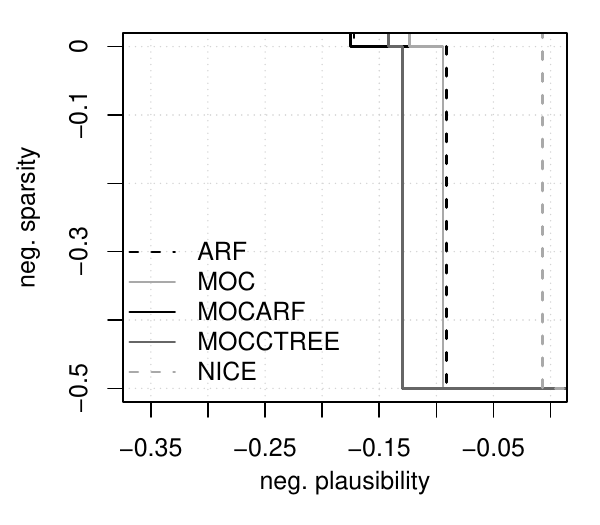}
  \vspace{-7mm}
  \caption{\texttt{two\_sines}}
  \label{fig:3}
\end{subfigure}

\medskip
\begin{subfigure}{0.3\textwidth}
  \includegraphics[width=\linewidth]{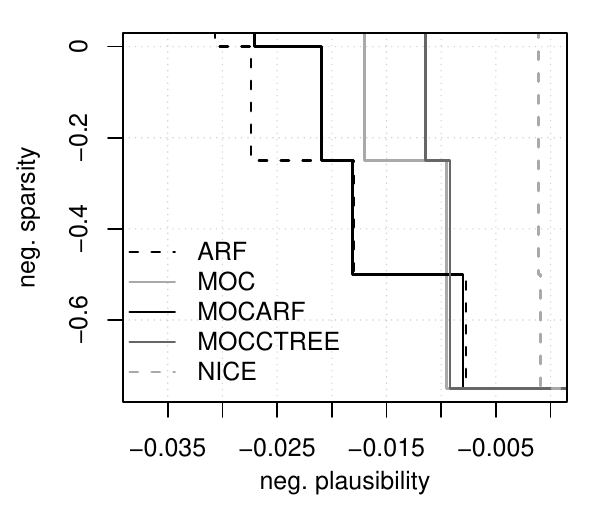}
  \vspace{-7mm}
  \caption{\texttt{bn\_5}}
  \label{fig:4}
\end{subfigure}\hfil %
\begin{subfigure}{0.3\textwidth}
  \includegraphics[width=\linewidth]{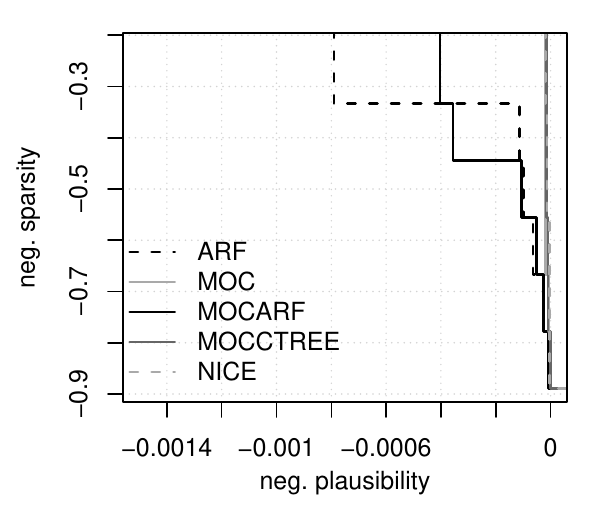}
  \vspace{-7mm}
  \caption{\texttt{bn\_10}}
  \label{fig:5}
\end{subfigure}\hfil %
\begin{subfigure}{0.3\textwidth}
  \includegraphics[width=\linewidth]{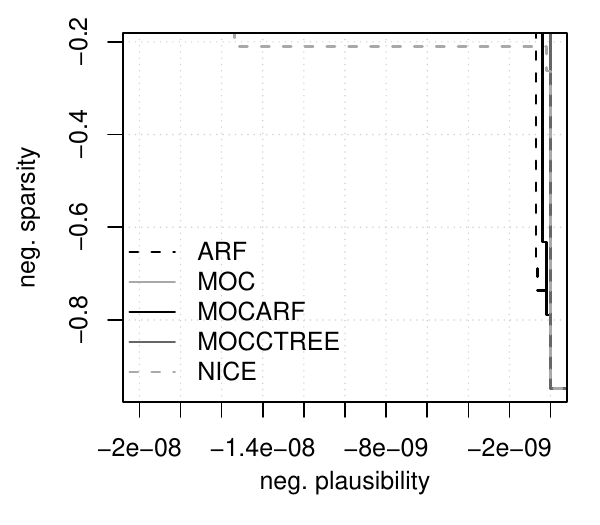}
  \vspace{-7mm}
  \caption{\texttt{bn\_20}}
  \label{fig:6}
\end{subfigure}
\caption{Median empirical attainment function \citep{lopez2010} for the negative plausibility and negative sparsity. Lower values are better.}
\label{fig:eaf-spars}
\end{figure}

\end{document}